\documentclass[runningheads]{llncs}

\usepackage[year=2026,ID={HCV}]{eccv}

\usepackage{eccvabbrv}
\usepackage{graphicx}
\usepackage{booktabs}
\usepackage{multirow}
\usepackage[nolist]{acronym}
\usepackage{float}
\usepackage{bbm}
\usepackage[accsupp]{axessibility}
\usepackage[breaklinks,colorlinks,citecolor=eccvblue]{hyperref}
\usepackage{wrapfig}
\usepackage{subcaption}
\usepackage{placeins}

\usepackage[expansion=false]{microtype}

\begin{document}

\setlength{\textfloatsep}{5pt plus 1pt minus 1pt}
\setlength{\floatsep}{5pt plus 1pt minus 1pt}
\setlength{\intextsep}{5pt plus 1pt minus 1pt}
\setlength{\abovecaptionskip}{3pt}
\setlength{\belowcaptionskip}{0pt}

\begin{acronym}[DRAM]
  \acro{IoR}{Inhibition of Return}
  \acro{WTA}{Winner-Take-All}
  \acro{RAM}{Recurrent Attention Model}
  \acro{DRAM}{Deep Recurrent Attention Model}
  \acro{GAP}{Glimpse-based Active Perception}
  \acro{FLOPs}{Floating Point Operations}
  \acro{DNF}{Dynamic Neural Field}
  \acro{mIoU}{mean Intersection over Union}
  \acro{CNN}{Convolutional Neural Network}
\end{acronym}

\title{Efficient Semantic Understanding from Digital Foveation}
\titlerunning{Semantic Understanding from Digital Foveation}


\author{
Caterina Caccavella\inst{1,2} \and
Vittorio Fra\inst{1,3} \and
Andreas Ziegler\inst{4} \and
Giulia D'Angelo\inst{5} \and
Yulia Sandamirskaya\inst{1}
}
\authorrunning{Caccavella et al.}

\institute{\small
Zurich University of Applied Sciences (ZHAW), W\"adenswil, Switzerland \and
ETH Z\"urich, Z\"urich, Switzerland \and
Politecnico di Torino, Turin, Italy \and
University of T\"ubingen, T\"ubingen, Germany \and
Czech Technical University, Prague, Czech Republic
}

\maketitle

\begingroup
\renewcommand\thefootnote{}
\footnotetext{The source code is available at \href{https://github.com/neuromorphic-zhaw/active-semantic-understanding}{github.com/neuromorphic-zhaw/active-semantic-understanding}}
\endgroup
\vspace{-0.5em}
\begin{abstract}
Dense semantic segmentation allocates computational resources uniformly across the entire image, regardless of scene complexity or task relevance. Inspired by biological vision, we investigate whether semantic understanding can be achieved more efficiently through digital foveated perception. We introduce a lightweight active-vision pipeline that combines saliency-driven fixation selection, high-resolution foveal observations, low-resolution contextual information, semantic accumulation, and adaptive computation. Beyond conventional dense prediction metrics, we use object-level evaluation to measure semantic understanding under sparse observations. On ADE20K-Object, a single foveated observation achieves 95.9\% of the baseline Top-1 accuracy and 96.9\% of the baseline Top-3 accuracy while requiring only 4.7\% of the computational cost. At the scene level, semantic accumulation recovers 90.6\% of the baseline object recall while using 58.6\% of the computation. These results suggest that substantial semantic understanding can emerge from sparse observations when computation is allocated selectively, highlighting active vision as an efficient alternative to uniform dense processing and motivating evaluation protocols beyond conventional pixel-wise segmentation metrics.

\end{abstract}

\section{Introduction}
\vspace{-0.5em}
\label{sec:intro}
Humans extract useful semantic information while processing only a fraction of the available visual input~\cite{ort2019humans}. Human vision achieves this through selective allocation of computational resources across space and time, most notably through visual attention and foveation, where high-resolution information is acquired only at selected locations while the visual periphery provides coarse contextual information~\cite{itti2001computational,martinez2004role,felleman1991distributed}. The process of directing relevant scene regions to the fovea through eye movements is known as \emph{foveation}~\cite{martinez2004role}, and can be approximated in artificial systems through \emph{digital foveation}, where high-resolution processing is restricted to selected regions of the visual field~\cite{mnih2014recurrent}.

Modern computer vision systems typically process images densely and uniformly, regardless of the scene structure or task requirements~\cite{guo2022attention}. While this strategy has enabled remarkable advances in recognition, detection, and segmentation, it also leads to rapidly increasing computational and energy requirements as image resolutions and model capacities continue to grow~\cite{strubell2019energy}. This trend is largely motivated by conventional imaging sensors, which provide dense images at a fixed spatial resolution~\cite{szeliski2022computer}, and by the efficiency of highly parallel GPU computation on dense tensors~\cite{raina2009large}. However, these assumptions become restrictive in robotics, autonomous systems, and embedded applications operating under strict resource constraints~\cite{chen2018constraint}.

This observation raises a fundamental twofold question: how much semantic information is actually required to understand a scene? And can this information be acquired more efficiently through selective processing? In this work, we investigate how semantic understanding emerges from sparse observations acquired through digital foveation. We introduce a biologically inspired active-vision pipeline that combines saliency-driven fixation selection, high-resolution foveal observations, low-resolution contextual information, semantic accumulation, and adaptive computation. By allocating computation to informative regions of a scene, the proposed pipeline investigates the relationship between semantic understanding, dense reconstruction, and computational cost. We adopt semantic segmentation as our learning and evaluation framework, since it provides a rich supervision signal that simultaneously encodes object identity, location, and extent, making it particularly well suited for learning semantic representations from sparse observations. However, dense metrics alone may not fully capture the semantic information acquired through active vision. We therefore complement conventional segmentation metrics with object-level measures of semantic understanding and scene-level evidence accumulation.

We evaluate the proposed pipeline on both controlled and realistic datasets, showing that substantial semantic understanding can emerge from sparse observations while requiring substantially less computation than full-image processing.

The \textbf{main contributions} of this work are as follows:
\vspace{-1.2em}
\begin{itemize}
\item We build a modular and interpretable active-vision pipeline based on biologically inspired, training-free attention mechanisms, including saliency, inhibition-of-return, contextual processing, and semantic accumulation.
\item We propose an evaluation protocol that complements conventional pixel-wise segmentation metrics with object-level measures of semantic understanding.
\item We show that active visual systems can acquire semantic information from sparse observations that approach the performance of full-image processing while requiring substantially less computation.
\item We explicitly characterize the trade-off between semantic understanding and computational cost on both controlled (CLEVR) and realistic (ADE20K) environments.
\end{itemize}
\vspace{-0.3em}

\section{Related Work}
\vspace{-0.5em}
Early computational models of visual attention were strongly influenced by findings in neuroscience and cognitive science. The seminal work of Itti et al.~\cite{itti2001computational} introduced a saliency-based model of bottom-up attention in which salient regions emerge through comparisons between local image features and their surrounding neighborhood across multiple feature channels and are sequentially selected through \ac{WTA} and \ac{IoR} mechanisms. Subsequent work extended these ideas through perceptual grouping~\cite{russell2014model}, event-based sensing~\cite{iacono2019proto}, depth-aware attention~\cite{ghosh2022event}, and low-power neuromorphic implementations~\cite{d2022event}. Moreover, biologically inspired frameworks such as \ac{DNF} have explored interpretable and modular cognitive architectures that integrate visual attention for perception and action~\cite{grieben2024roboverine}.

Attention mechanisms were later incorporated into artificial vision systems through glimpse-based models such as the Recurrent Attention Model (RAM)~\cite{mnih2014recurrent} and Deep RAM~\cite{ba2014multiple}. More recent active-vision frameworks include Glimpse-based Active Perception and Coarse-to-Fine GAP~\cite{kolner2025mind,kolner2025task}, which combine saliency-driven attention with modern neural architectures, as well as Active Visual Exploration Based on Attention-Map Entropy (AME)~\cite{pardyl2023active}, AdaptiveNN~\cite{wang2025emulating}, and the Canvas Vision Transformer (CanViT)~\cite{berreby2026canvit}, which investigate attention-guided exploration, adaptive computation, and scene-level memory representations. Collectively, these works demonstrate the effectiveness of sparse visual observations, although they are typically optimized end-to-end for a specific downstream objective, such as classification, detection, or dense prediction. Recent work has also explored biologically inspired processing closer to the sensor level. The Foveated Vision Interface (FOVI)~\cite{blauchbiologically} introduces a retina-inspired sampling interface that substantially reduces computation for image classification, although it does not yet address active fixation selection.

In contrast, our goal is to investigate how semantic information is progressively acquired and integrated from sparse observations and how this process relates to computational efficiency. We therefore focus on a lightweight segmentation model combined with a training-free attention mechanism suitable for resource-constrained settings, while systematically evaluating how semantic evidence is acquired and accumulated through sequential glimpses.

\vspace{-0.5em}
\section{Methods}
\vspace{-0.3em}

We study active visual understanding through a biologically inspired digital foveation pipeline. Informative scene regions are identified using a low-resolution saliency mechanism and explored through sequential high-resolution foveal observations augmented with low-resolution context. Semantic information acquired across fixations is accumulated in a scene-level memory representation. A schematic of the proposed pipeline is shown in~\cref{fig:overview}, and additional implementation details are provided in the supplementary material.

\begin{figure}[!t]
  \centering
  \includegraphics[width=\linewidth]{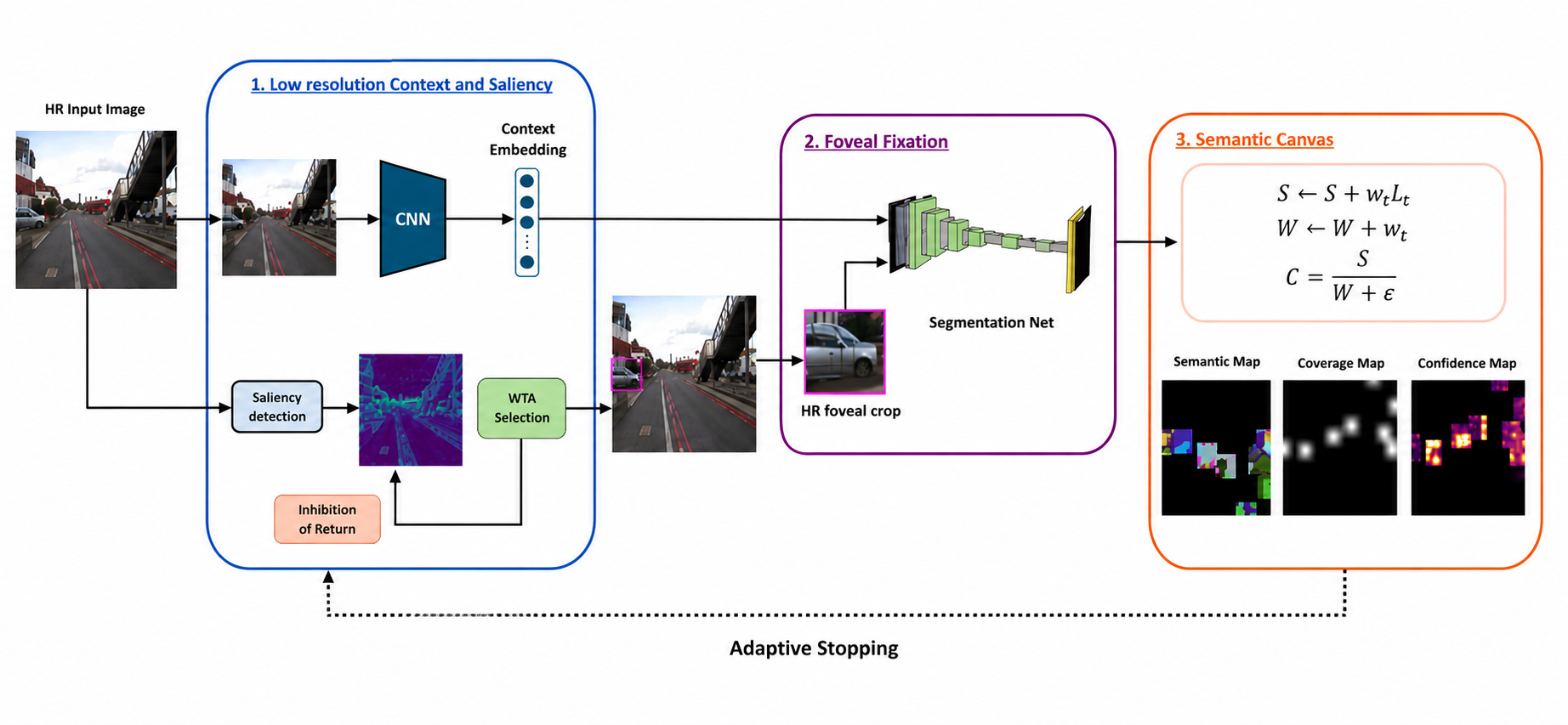}
    \caption{Overview of the proposed active-vision semantic segmentation pipeline. A low-resolution preview provides saliency and global context, while a \acf{WTA} mechanism with inhibition-of-return selects fixation locations. At each fixation, a high-resolution foveal crop is segmented and its logits are accumulated in a semantic canvas to produce the final prediction.}
  \label{fig:overview}
\end{figure}

\noindent\textbf{Digital Foveation and Fixation Selection.}
A low-resolution copy of the input image is used both to compute a saliency map, as in~\cite{itti2001computational}, and to provide global scene context. Fixations are selected from this saliency map using inhibition of return (IoR), which suppresses previously attended locations and encourages exploration without training a separate policy. 
At each fixation, the model extracts a high-resolution foveal crop, while the low-resolution full image is processed once by a lightweight context CNN comprising three convolutional blocks, global average pooling, and a linear projection to produce a compact context embedding.
The resulting context embedding is projected and added to the foveal network features, providing coarse scene-level information without requiring dense full-image segmentation.

\noindent\textbf{Training from Sparse Object Views.}
During training, fixation locations are sampled from semantic object masks rather than saliency maps. Saliency naturally favors large and visually distinctive regions and would therefore produce an imbalanced training distribution. Since saliency serves primarily as a training-free bottom-up mechanism for selecting observations at inference time, we decouple the training distribution from the saliency distribution and instead sample object categories more uniformly.
For each fixation, a high-resolution foveal crop and a low-resolution global context are extracted. Supervision is applied to all valid semantic pixels within the crop, allowing the model to learn object representations from partial observations acquired at different viewpoints and spatial locations while preserving standard dense segmentation supervision on the observed region.

\noindent\textbf{Semantic Canvas and Adaptive Computation.}
The semantic canvas is a scene-level memory that stores semantic predictions accumulated across multiple fixations. After each fixation, the local semantic logits predicted by the segmentation network are projected back into their corresponding image coordinates. Before being written into the canvas, each logit map is multiplied by a Gaussian weighting mask centered on the fixation, giving greater influence to predictions near the center of the foveal crop and smoothly reducing the contribution of predictions toward the crop boundaries. A second canvas stores the corresponding accumulated weights, allowing overlapping observations to be combined through weighted averaging and indicating which image regions have been observed. The final semantic prediction is obtained by normalizing the accumulated logits with the accumulated weights and selecting the most likely class at each pixel. This update mechanism is entirely deterministic and contains no trainable recurrent memory or learned write rule.
In resource-constrained settings, the number of fixations can be chosen according to the task or computational budget, while fixation selection is driven by saliency and inhibition of return. The same canvas also provides the basis for adaptive computation: confidence and coverage estimates can be used to stop once additional fixations are unlikely to add useful semantic evidence, allowing simple scenes to use fewer observations and more complex scenes to receive additional computation.

\vspace{-0.6em}
\section{Experimental Setup}
\vspace{-0.6em}
\noindent\textbf{Datasets.}
We evaluate the proposed pipeline in both controlled and realistic environments. As a controlled proof-of-principle benchmark, we use CLEVR~\cite{johnson2017clevr}, a synthetic dataset with well-defined object categories and boundaries. To evaluate active vision in realistic scenes, we construct an object-centric subset of ADE20K~\cite{zhou2017scene} containing 98 object classes by excluding layout and background categories. The complete class list is provided in the supplementary material.

\noindent\textbf{Baselines.}
We compare against the off-the-shelf lightweight semantic segmentation network Lite R-ASPP~\cite{howard2019searching}, which is publicly available and widely used for dense prediction tasks. The dense baseline processes the entire image in a single forward pass and serves as reference for both accuracy and computational cost.

\noindent\textbf{Metric.}
Computational efficiency is measured in \ac{FLOPs}. The active vision pipeline separates computation into a one-time per-image cost and a per-fixation cost. The one-time cost includes the low-resolution context branch and, when reported, the saliency computation. The per-fixation cost includes extracting the foveal crop, running the foveal segmentation network, and writing the predicted logits into the semantic canvas. Since the context embedding can be cached and reused across fixations, the total cost grows approximately linearly with the number of foveal observations. 

\noindent\textbf{Evaluation Protocol.}
We evaluate active vision from three complementary perspectives: dense reconstruction, semantic understanding, and computational efficiency. Dense reconstruction is measured using full-image \ac{mIoU} and scene coverage. Semantic understanding is assessed using object-level metrics including Top-1 and Top-3 accuracy, visible-object \ac{mIoU}, and object recall. Computational efficiency is measured in GFLOPs and reported relative to the corresponding full-image baseline.

\section{Results}
\vspace{-0.4em}
\subsection{Single-Fixation Semantic Understanding}
\vspace{-0.1em}
As shown in~\cref{tab:single_fixation_summary}, a single foveated observation captures a large fraction of the semantic information available to the full-image model while using substantially less computation. Although the glimpse model has slightly more parameters due to its modular architecture and additional context-processing components, its computational cost remains significantly lower because only small foveal regions are processed at high resolution.

On CLEVR, the full-image baseline processes $240{\times}240$ images, requires 1.033 GFLOPs, and reaches 0.929 Top-1 accuracy, 0.997 Top-3 accuracy, and 0.648 visible \ac{mIoU} on the object-level readout. The glimpse model processes $40{\times}40$ foveal crops and requires only 12.3\% of the baseline cost, while reaching 0.982 Top-1 accuracy, 1.000 Top-3 accuracy, and 0.961 visible \ac{mIoU}. This indicates that, in the controlled CLEVR setting, attended local observations are sufficient to recover object-level semantics reliably.

On ADE20K-Object, the full-image baseline processes $512{\times}512$ images and requires 4.361 GFLOPs, whereas the fovea plus context glimpse model processes a $96{\times}96$ foveal crop and requires only 4.7\% of the baseline cost. Evaluated on the same visible object pixels, the glimpse model reaches 0.509 Top-1 accuracy, 0.715 Top-3 accuracy, and 0.243 visible \ac{mIoU}, retaining 95.9\%, 96.9\%, and 72.3\% of the corresponding baseline performance, respectively.

\begin{table}[t]
\centering
\small
\renewcommand{\arraystretch}{0.92}
\setlength{\tabcolsep}{4pt}
\caption{Single-observation semantic understanding. The full-image baseline is evaluated on the same visible object pixels as the foveated model.}
\label{tab:single_fixation_summary}
\begin{tabular}{@{}llcccccc@{}}
\toprule
Dataset & Model & Params & GFLOPs($\downarrow$)  & Top-1($\uparrow$) & Top-3($\uparrow$) & Visible($\uparrow$) \\
\midrule
\multirow{2}{*}{CLEVR} & Baseline & 3.962M & 1.033 & 0.929 & 0.997 & 0.648 \\
 & Glimpse-model & 4.185M & 0.127 & 0.982 & 1.000 & 0.961 \\
\midrule
\multirow{2}{*}{\shortstack{ADE20K\\Obj}}& Baseline & 3.493M & 4.361 & 0.531 & 0.738 & 0.336 \\
 & Glimpse-model & 3.687M & 0.206 & 0.509 & 0.715 & 0.243 \\
\bottomrule
\end{tabular}
\end{table}

\vspace{-0.3em}
\subsection{Learning Semantic Representations from Sparse Observations}
\vspace{-0.1em}
\begin{table}[b]
\centering
\small
\caption{Effect of use of contextual information on ADE20K-Object. All models are evaluated using the same single-fixation protocol on visible object pixels.}
\label{tab:training_ablation}
\begin{tabular}{lccc}
\toprule
Method &
Top-1($\uparrow$) &
Top-3($\uparrow$) &
Visible \ac{mIoU}($\uparrow$) \\
\midrule
Full-object crop training & 0.385 & 0.593 & 0.195 \\
Fovea only & 0.499 & 0.709 & 0.257 \\
Fovea + context & 0.509 & 0.715 & 0.261 \\
\bottomrule
\end{tabular}
\end{table}

\Cref{tab:training_ablation} evaluates the role of contextual information and sparse-view training. Adding a low-resolution global context consistently improves performance, increasing Top-1 accuracy from 0.499 to 0.509, Top-3 accuracy from 0.709 to 0.715, and visible-object \ac{mIoU} from 0.257 to 0.261. These gains are most pronounced for categories such as beds, desks, countertops, and armchairs, where local appearance alone may be insufficient for reliable recognition. This observation is consistent with previous studies showing that modern vision networks rely on non-local contextual information and exhibit effective receptive fields that extend well beyond local image regions~\cite{luo2016understanding,wong2020targeted}. Additional per-class results are provided in the supplementary material. We further compare sparse-view training against a full-object crop control, where each training sample consists of a crop containing the entire object.
Despite having access to complete object views during training, the full-object model reaches only 0.385 Top-1 accuracy and 0.195 visible-object \ac{mIoU}, substantially below the sparse-view model. This suggests that learning from partial observations acquired at different viewpoints and spatial locations is not merely a computational constraint, but can produce representations that transfer more effectively to foveated evaluation.

\vspace{-0.3em}
\subsection{Scene Understanding Through Sequential Semantic Integration}
\vspace{-0.1em}

\begin{wrapfigure}{r}{0.42\linewidth}
    \centering
    \vspace{-0.6em}
    \includegraphics[width=\linewidth]{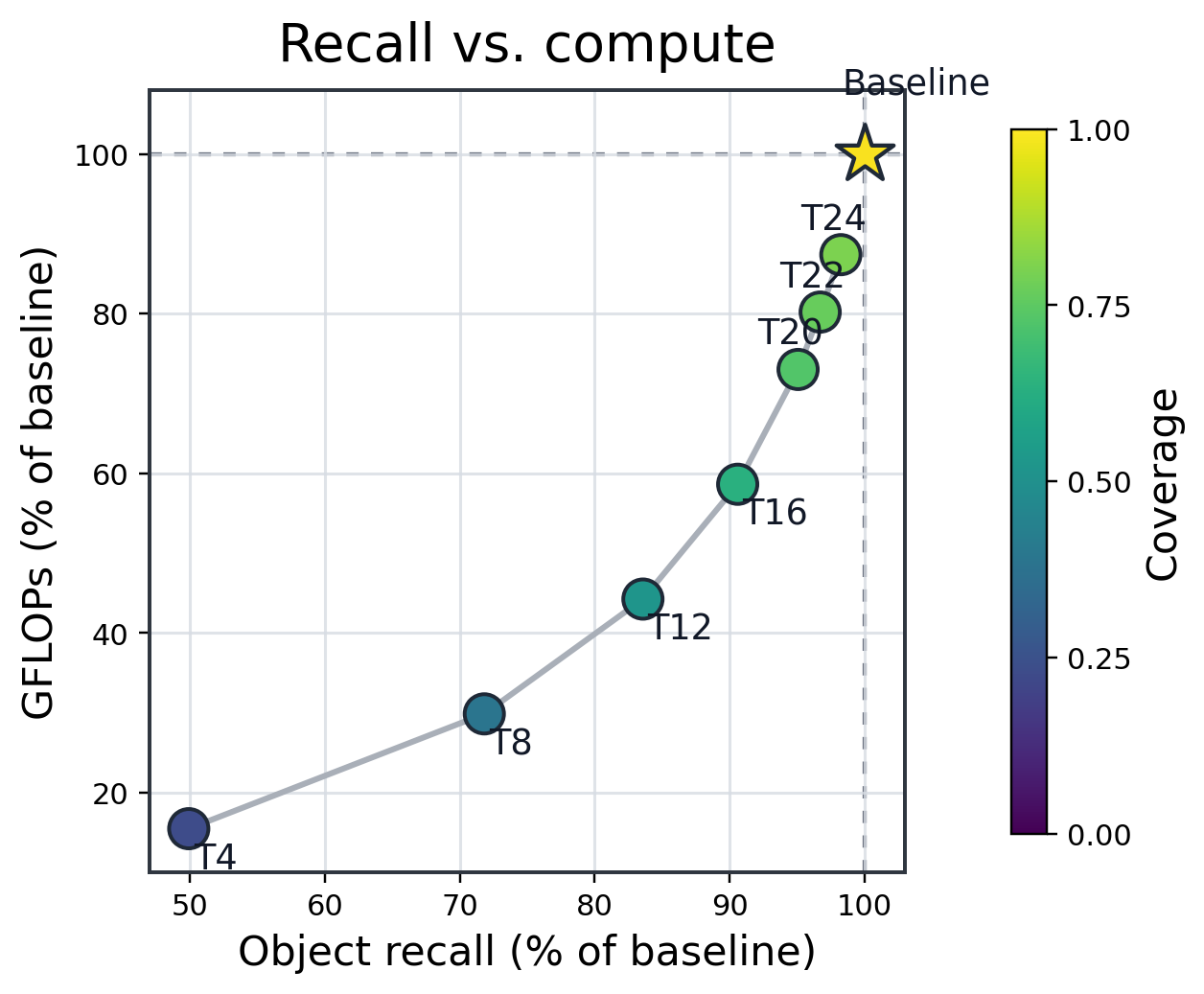}
    \caption{Object recall vs. inference FLOPs on ADE20K-Object. Color indicates canvas coverage; each point is a fixed fixation budget.}
    \label{fig:recall_vs_flops}
    \vspace{-1em}
\end{wrapfigure}

Multiple observations allow the model to accumulate semantic evidence over time. We first verify this behavior on CLEVR using sparse foveal observations.

Using 4 fixations, the model covers 80.4\% of objects (6.38 per scene on average) and reaches 0.391 full-image \ac{mIoU}, 0.862 pixel accuracy, 0.900 Top-1, and 0.996 Top-3 accuracy over observed regions. Increasing the budget to 6 fixations improves object coverage to 90.7\%, demonstrating that object-level semantic information can be recovered from only a few targeted observations.

\Cref{fig:recall_vs_flops} reports the corresponding evidence-accumulation behavior on ADE20K-Object. Here, the trade-off between computation, object discovery, and dense reconstruction is more pronounced. Object recall increases with the number of fixations $T$, reaching 0.645 at $T=16$, corresponding to 90.6\% of the full-image baseline recall while requiring only 58.6\% of the baseline computation. Top-1 accuracy reaches 93.7\% of the full-image baseline. Increasing the budget to $T=24$ further improves object recall to 98.2\% of the baseline while using 87.4\% of the baseline computation. However, full-canvas \ac{mIoU} remains below the baseline, indicating that sparse observations recover object semantics faster than dense scene reconstruction. The complete set of results discussed here are reported in the supplementary material.

Additional experiments on adaptive stopping and goal-directed visual search are also reported in the supplementary material.

\vspace{-0.3em}
\section{Discussion}
\vspace{-0.3em}
Our results suggest that useful semantic understanding can emerge from sparse observations without requiring uniform processing of the entire visual scene. Across both controlled and realistic environments, object-level recognition performance remains close to that of full-image processing despite substantial reductions in computational cost. However, dense reconstruction remains more challenging, suggesting that conventional segmentation metrics may underestimate the semantic information captured by active vision. These findings support a broader shift in perspective: rather than focusing exclusively on matching state-of-the-art dense prediction benchmarks, active vision research should increasingly investigate how semantic representations are acquired, accumulated, and utilized under realistic computational constraints.

More generally, we argue that progress toward efficient and adaptive visual systems may require rethinking not only learning algorithms but also the way visual information is sampled, represented, and processed. Modern vision systems continue to heavily rely on scaling model size and training data to achieve generalization. In contrast, biological systems appear to build reusable representations from sparse observations through hierarchical, modular, and highly adaptive mechanisms. While our study explores only one instance of this idea through digital foveation, related efforts can be found across neuroscience, neuromorphic sensing, active perception, and cognitive architectures~\cite{schoner2016dynamic,friston2017active,d2025wandering}. 

The present work nevertheless has some limitations. First, our evaluation is restricted to CLEVR and ADE20K-Object, and therefore does not extensively capture the complexity of real-world embodied perception. Second, our approach relies on digital foveation rather than physical sensing systems and camera motion. Finally, the semantic canvas currently acts primarily as a memory mechanism and only weakly influences fixation selection and stopping decisions. Future work should investigate richer scene representations capable of actively guiding exploration, adaptive stopping, and task-dependent behavior. A natural extension is to integrate top-down knowledge into the semantic canvas, enabling memory and task goals to guide future observations. 

\vspace{-0.6em}
\section{Conclusion}
\vspace{-0.7em}
We presented a biologically inspired active-vision framework that combines saliency-driven fixation selection, digital foveation, semantic accumulation, and adaptive computation. At the object level, a single foveated observation achieves 95.9\% of the baseline Top-1 accuracy and 96.9\% of the baseline Top-3 accuracy while requiring only 4.7\% of the computation. At the scene level, progressively integrating semantic information across 12 fixations recovers 83.6\% of the baseline object recall while using 44.3\% of the baseline computation, whereas increasing the budget to 16 fixations improves object recall to 90.6\% of the baseline while using 58.6\% of the computation. These results demonstrate that semantic understanding can emerge from sparse observations and that the number of fixations provides a simple and controllable trade-off between semantic performance and computational cost. This provides a promising foundation for future active-vision systems incorporating richer semantic memory, top-down reasoning, and task-driven exploration.

\section{Acknowledgment}
This work was supported by the SNSF Practice to Science grant PT0042\_222692 „Brain-inspired vision technologies for assistive robots", the Fluently project with Grant Agreement No. 101058680 and the European Union - NextGenerationEU Project 3A-ITALY MICS (PE0000004, CUP E13C22001900001, Spoke 6).

\bibliographystyle{splncs04}
\bibliography{main}

\clearpage

\title{Supplementary Material}
\author{}
\institute{}
\maketitle

\appendix
\renewcommand{\theHsection}{A\arabic{section}}

\setcounter{figure}{0}
\setcounter{table}{0}
\renewcommand{\thefigure}{S\arabic{figure}}
\renewcommand{\thetable}{S\arabic{table}}

\section{Additional Implementation Details}

\subsection{Semantic Canvas}
The semantic canvas provides an explicit memory of the semantic evidence collected across fixations. At fixation $t$, the foveal network predicts a local semantic logit map $L_t$ that is projected back into image coordinates. To account for the varying reliability of predictions within the foveal crop, each logit map is weighted by a Gaussian weighting mask $w_t$, centered on the fixation location, assigning higher weight to pixels near the fixation center and lower weight to pixels near crop boundaries. The canvas stores both accumulated logits $S$ and accumulated write weights $W$:

\begin{equation}
S \leftarrow S + w_t L_t,
\qquad
W \leftarrow W + w_t.
\end{equation}

After $T$ fixations, the final semantic canvas is obtained by normalizing the accumulated logits,
\begin{equation}
C = \frac{S}{W+\epsilon},
\end{equation}
where $\epsilon$ is a small constant used for numerical stability. Semantic predictions are obtained by taking the class with maximum logit value at each pixel.

The accumulated weight map $W$ additionally provides a measure of scene coverage, indicating which regions have received semantic evidence. Unlike recurrent memory architectures, the semantic canvas contains no learned update mechanism; it simply accumulates semantic evidence directly in image coordinates through weighted averaging.
\vspace{-0.7em}
\subsubsection{Additional Semantic Canvas Ablations}
\vspace{-0.5em}
We explored several mechanisms for integrating semantic evidence across fixations. These ablations were not used as the main result because they did not provide the best overall accuracy--efficiency trade-off, but they highlight the flexibility of the proposed modular framework and motivate future work.

\textbf{\textit{Gaussian logit accumulation.}}
The main model uses a non-parametric semantic canvas in which foveal logits are written back to image coordinates using Gaussian-weighted averaging. This design is inexpensive, stable, and modular: the foveal recognizer, saliency policy, and canvas can be modified independently. It provides the best trade-off when the goal is efficient object discovery and partial scene understanding.

\textbf{\textit{Low-resolution canvas priors.}}
We also tested initializing the canvas with a weak low-resolution semantic prior computed from the global preview image. This improved early scene coverage and full-canvas \ac{mIoU}, especially at low fixation budgets, but introduced additional computation and did not improve object-level Top-1 accuracy. This variant is useful when broad scene completion is more important than minimizing computation.

\textbf{\textit{Learned scene-belief canvas.}}
A stronger variant used a learned stateful belief canvas initialized from a low-resolution semantic prior and refined after each fixation using the accumulated foveal evidence. This substantially improved dense scene completion and object discovery: at $T=20$, full-canvas \ac{mIoU} increased from 0.184 to 0.220, and object recall increased from 0.677 to 0.983. However, Top-1 object accuracy decreased from 0.545 to 0.500 and computational cost increased. We therefore treat it as a higher-compute variant for applications requiring denser scene maps.


\textbf{\textit{Multi-scale glimpses.}}
Earlier experiments used multi-scale glimpse heads that predicted semantic maps at multiple spatial scales, taking inspiration from the multi-scale glimpse sensor from \hyperlink{cite.mnih2014recurrent_supp}{\textcolor{eccvblue}{[1]}}. These provided broader context and greater flexibility, but dense supervision of peripheral predictions often biased the representation toward coarse or dominant regions. The final architecture instead uses low-resolution context as a feature cue while keeping the main semantic prediction foveal. \newline

Overall, these ablations show that the proposed framework can be adapted to different task requirements. For object discovery or efficient partial mapping, the observed-evidence Gaussian canvas is preferable. For denser semantic completion, low-resolution priors or learned scene-belief canvases improve coverage and full-canvas \ac{mIoU} at higher cost. For visual search, the same canvas can be queried or biased toward task-relevant classes without changing the foveal recognizer.

\subsection{Adaptive Stopping}
The adaptive stopping mechanism uses the semantic canvas to determine whether additional fixations are likely to provide useful information. No policy is learned; the mechanism operates entirely at inference time by combining bottom-up saliency, inhibition of return, and semantic confidence.

At fixation step $t$, candidate locations are scored according to
\begin{equation}
F_t(p)
=
\left(
A(p)
+
\alpha U_t(p)
\right)
I_t(p),
\end{equation}
where $A(p)$ denotes the saliency map, $U_t(p)$ the canvas uncertainty, and $I_t(p)$ the inhibition-of-return mask. Salient regions with high semantic uncertainty are therefore prioritized, while previously visited locations are suppressed.

Stopping is based on the remaining unexplained evidence,
\begin{equation}
R_t(p) = A(p) I_t(p) U_t(p).
\end{equation}
Exploration terminates when the maximum remaining score, $\max_p R_t(p)$, stays below a threshold $\tau$ for two consecutive fixations after a minimum fixation budget has been reached. Different computation--accuracy trade-offs are obtained by varying the uncertainty weight $\alpha$, the stopping threshold $\tau$, and the minimum fixation budget. No additional training is required. Preliminary results are reported in \ref{sec:ada_stopping_results}.
\subsection{Computational Cost}
\vspace{-0.8em}
Computational cost is measured in floating-point operations (FLOPs) used at the inference stage. We report FLOPs relative to the corresponding full-image baseline for each dataset, since CLEVR and ADE20K use different input resolutions.

For the active model, computation is decomposed into one-time per-image costs and per-fixation costs:
\begin{equation}
\mathrm{FLOPs}_{\mathrm{active}}(T)
=
\mathrm{FLOPs}_{\mathrm{fixed}}
+
T \cdot \mathrm{FLOPs}_{\mathrm{fovea}},
\end{equation}
where $\mathrm{FLOPs}_{\mathrm{fixed}}$ includes saliency estimation, contextual processing, and optional low-resolution prior computation, while $T$ denotes the number of fixations. For adaptive stopping experiments, $T$ is replaced by the average number of fixations used before termination.

The saliency map and low-resolution context embedding are computed once per image and reused across fixations. Canvas accumulation, inhibition of return, cropping, and bookkeeping operations are deterministic tensor operations and are small compared with the neural network forward passes. Including these operations does not materially change the reported trade-offs.

\section{ADE20K-Object Class Definition}
To construct ADE20K-Object, we retain semantic categories corresponding primarily to localized objects and assign all remaining stuff, layout, and structural categories to the ignore label during training and evaluation. Retained and excluded object classes are reported in~\cref{tab:kept_classes,tab:excluded_classes}.

\begin{table}[!t]
\centering
\caption{Semantic object classes retained in ADE20K-Object.}
\label{tab:kept_classes}
\footnotesize
\begin{tabular}{ll|ll|ll}
\toprule
ID & Class & ID & Class & ID & Class \\
\midrule
7 & bed & 55 & case & 107 & washer \\
8 & windowpane & 56 & pool table & 108 & plaything \\
10 & cabinet & 57 & pillow & 110 & stool \\
12 & person & 58 & screen door & 111 & barrel \\
14 & door & 62 & bookcase & 112 & basket \\
15 & table & 63 & blind & 115 & bag \\
17 & plant & 64 & coffee table & 116 & minibike \\
18 & curtain & 65 & toilet & 117 & cradle \\
19 & chair & 66 & flower & 118 & oven \\
20 & car & 67 & book & 119 & ball \\
22 & painting & 69 & bench & 120 & food \\
23 & sofa & 70 & countertop & 124 & microwave \\
24 & shelf & 71 & stove & 125 & pot \\
27 & mirror & 73 & kitchen island & 126 & animal \\
28 & rug & 74 & computer & 127 & bicycle \\
30 & armchair & 75 & swivel chair & 129 & dishwasher \\
31 & seat & 76 & boat & 130 & screen \\
33 & desk & 77 & bar & 131 & blanket \\
35 & wardrobe & 78 & arcade machine & 132 & sculpture \\
36 & lamp & 80 & bus & 133 & hood \\
37 & bathtub & 81 & towel & 134 & sconce \\
39 & cushion & 82 & light & 135 & vase \\
41 & box & 83 & truck & 136 & traffic light \\
43 & signboard & 85 & chandelier & 137 & tray \\
44 & chest of drawers & 87 & streetlight & 138 & ashcan \\
45 & counter & 88 & booth & 139 & fan \\
47 & sink & 89 & television & 141 & crt screen \\
49 & fireplace & 90 & airplane & 142 & plate \\
50 & refrigerator & 92 & apparel & 143 & monitor \\
&&97 & ottoman & 145 & shower \\
&&98 & bottle & 146 & radiator \\
&&99 & buffet & 147 & glass \\
&&100 & poster & 148 & clock \\
&&102 & van & 149 & flag \\
&&103 & ship && \\
\bottomrule
\end{tabular}
\end{table}

\begin{table}[!t] \centering \caption{Background and layout classes excluded from ADE20K-Object.} \label{tab:excluded_classes} \footnotesize \begin{tabular}{ll|ll|ll} \toprule ID & Class & ID & Class & ID & Class \\ \midrule 0 & wall & 48 & skyscraper & 101 & stage \\ 1 & building & 51 & grandstand & 104 & fountain \\ 2 & sky & 52 & path & 105 & conveyer belt \\ 3 & floor & 53 & stairs & 106 & canopy \\ 4 & tree & 54 & runway & 109 & swimming pool \\ 5 & ceiling & 59 & stairway & 113 & waterfall \\ 6 & road & 60 & river & 114 & tent \\ 9 & grass & 61 & bridge & 121 & step \\ 11 & sidewalk & 68 & hill & 122 & storage tank \\ 13 & earth & 72 & palm & 123 & trade name \\ 16 & mountain & 79 & hovel & 128 & lake \\ 21 & water & 84 & tower & 140 & pier \\ 25 & house & 86 & awning & 144 & bulletin board \\ 26 & sea & 91 & dirt track && \\ 29 & field & 93 & pole && \\ 32 & fence & 94 & land && \\ 34 & rock & 95 & bannister && \\ 38 & railing & 96 & escalator && \\ 40 & base &&&& \\ 42 & column &&&& \\ 46 & sand &&&& \\ \bottomrule \end{tabular} \end{table}
\section{Additional Experiments}

This section reports additional experiments that complement the main paper, including adaptive stopping, goal-directed visual search, and per-class analyses.

\subsection{Per-Class Analysis}

To better understand the role of context and sparse observations, we analyze performance at the level of individual object categories. Two complementary questions are considered: (i) which classes benefit most from low-resolution contextual information, and (ii) which classes can be reliably recognized from sparse observations.

\subsubsection{Classes Benefiting from Context}

\Cref{tab:context_gains} reports the object classes exhibiting the largest improvements when low-resolution context is added to the foveal observation. The strongest gains are observed for categories whose appearance is strongly tied to the surrounding scene, including beds, desks, countertops, and armchairs. Context also improves recognition of visually ambiguous categories such as plates and ashcans, where local appearance alone is often insufficient for reliable classification.

These results support the hypothesis that coarse scene-level information provides useful semantic cues even when detailed recognition is performed locally through foveal observations.

\begin{table}[h]
\centering
\caption{Object classes with the largest gains from low-resolution context on ADE20K-Object.}
\label{tab:context_gains}
\begin{tabular}{lccccc}
\toprule
Class & $n$ & Fovea Top-1 & Context Top-1 & $\Delta$ Top-1 & $\Delta$ mIoU \\
\midrule
Ashcan & 300 & 0.283 & 0.377 & +0.093 & +0.028 \\
Plate & 165 & 0.164 & 0.255 & +0.091 & +0.050 \\
Bed & 567 & 0.628 & 0.713 & +0.085 & +0.086 \\
Ship & 12 & 0.000 & 0.083 & +0.083 & +0.131 \\
Desk & 183 & 0.273 & 0.355 & +0.082 & +0.056 \\
Traffic light & 195 & 0.210 & 0.287 & +0.077 & +0.014 \\
Kitchen island & 27 & 0.148 & 0.222 & +0.074 & +0.045 \\
Sconce & 324 & 0.432 & 0.500 & +0.068 & +0.100 \\
Countertop & 93 & 0.290 & 0.355 & +0.065 & +0.073 \\
Armchair & 294 & 0.248 & 0.306 & +0.058 & +0.076 \\
\bottomrule
\end{tabular}
\end{table}

\subsubsection{Per-Class Recognition Performance}

Performance varies substantially across semantic categories. Classes with distinctive local visual evidence, such as car, plant, person, bed, curtain, and windowpane, are recognized reliably from sparse observations. For example, the glimpse model reaches Top-1 accuracies of 0.810 for car, 0.806 for plant, 0.758 for person, 0.713 for bed, and 0.708 for curtain.

In contrast, categories such as wardrobe, seat, counter, cabinet, sofa, and table remain considerably more challenging. These classes often exhibit higher appearance variability, partial visibility, and stronger dependence on contextual information, resulting in a larger performance gap relative to the full-image baseline. Representative per-class results are reported in~\cref{tab:per_class_object}.

\begin{table}[h]
\centering
\caption{Representative object-level performance by class on ADE20K-Object. The full-image baseline is evaluated on the same visible object pixels as the glimpse model.}
\label{tab:per_class_object}
\resizebox{\linewidth}{!}{
\begin{tabular}{lcccc}
\toprule
Class & Glimpse mIoU & Baseline mIoU & Glimpse Top-1 & Baseline Top-1 \\
\midrule
Bed & 0.6696 & 0.8412 & 0.7125 & 0.8713 \\
Windowpane & 0.6764 & 0.7198 & 0.6920 & 0.7021 \\
Cabinet & 0.3844 & 0.5002 & 0.3724 & 0.4609 \\
Person & 0.7711 & 0.8457 & 0.7581 & 0.8001 \\
Door & 0.4453 & 0.5672 & 0.4798 & 0.5400 \\
Table & 0.4184 & 0.5539 & 0.4911 & 0.5821 \\
Plant & 0.8592 & 0.8607 & 0.8060 & 0.7903 \\
Curtain & 0.6682 & 0.7177 & 0.7080 & 0.6972 \\
Chair & 0.4485 & 0.5522 & 0.5167 & 0.5615 \\
Car & 0.8371 & 0.8659 & 0.8101 & 0.8058 \\
Sofa & 0.4094 & 0.5936 & 0.4528 & 0.6195 \\
Wardrobe & 0.1535 & 0.5723 & 0.1705 & 0.5504 \\
Seat & 0.2121 & 0.5853 & 0.1548 & 0.4345 \\
Counter & 0.1123 & 0.2920 & 0.1204 & 0.2130 \\
\bottomrule
\end{tabular}
}
\end{table}

\subsection{Sequential Semantic Integration}
Table~\ref{tab:canvas_tradeoff} reports how semantic evidence accumulates as the number of fixations increases. Each fixation contributes a foveal semantic prediction that is written into the global canvas using Gaussian-weighted logit accumulation. FLOPs are computed with inference-only accounting: the low-resolution context is processed once per image and reused across fixations, while each additional fixation adds one cached-context foveal forward pass. As expected, increasing $T$ improves coverage, object recall, Top-1 accuracy, and full-canvas \ac{mIoU}, while increasing computation approximately linearly. At $T=16$, the model recovers 90.6\% of the full-image baseline object recall and 93.7\% of its Top-1 accuracy while using only 58.6\% of the baseline computation. Full-canvas \ac{mIoU} remains lower than the baseline, indicating that sparse foveal observations are more effective for accumulating object-level semantic evidence than for complete dense scene reconstruction.

\begin{table}[t]
\centering
\footnotesize
\caption{Semantic canvas performance on ADE20K-Object as a function of the number of fixations (T). FLOPs are computed using inference-only accounting: the low-resolution context is computed once per image and each fixation adds one foveal forward pass.}
\label{tab:canvas_tradeoff}
\begin{tabular}{lccccc}
\toprule
Method &
GFLOPs &
Coverage &
Object Recall &
Top-1 &
Full \ac{mIoU} \\
&
($\downarrow$) &
($\uparrow$) &
($\uparrow$) &
($\uparrow$) &
($\uparrow$) \\
\midrule
T4  & 0.676 & 0.227 & 0.355 & 0.485 & 0.068 \\
T8  & 1.303 & 0.387 & 0.511 & 0.512 & 0.111 \\
T12 & 1.930 & 0.521 & 0.595 & 0.524 & 0.141 \\
T16 & 2.557 & 0.634 & 0.645 & 0.537 & 0.165 \\
T20 & 3.183 & 0.727 & 0.677 & 0.545 & 0.184 \\
T22 & 3.497 & 0.766 & 0.688 & 0.548 & 0.192 \\
T24 & 3.810 & 0.801 & 0.699 & 0.553 & 0.198 \\
\midrule
Full-image baseline & 4.361 & 1.000 & 0.712 & 0.573 & 0.376 \\
\bottomrule
\end{tabular}
\end{table}

\subsection{Adaptive Stopping}
\label{sec:ada_stopping_results}

Adaptive stopping was evaluated on both CLEVR and ADE20K-Object. The mechanism is not learned and operates only at inference time. At each fixation, candidate locations are scored by combining bottom-up saliency, inhibition of return, and semantic uncertainty estimated from the current canvas. Exploration terminates when the remaining salient uncertainty stays below a threshold for several consecutive fixations after a minimum fixation budget has been reached.

On CLEVR, adaptive stopping was evaluated with a semantic canvas initialized by a low-resolution semantic prior and refined by foveal observations. The prior is a coarse full-image prediction computed once from a downsampled image and inserted into the canvas with low confidence. This provides an initial estimate of semantic confidence over the entire scene, while subsequent foveal fixations locally refine the canvas. With a low-compute setting, the model used 4.0 fixations on average, requiring 0.245 GFLOPs, or 23.7\% of the full-image baseline, while covering 80.4\% of the objects. Increasing the average number of fixations to 6.0 raised object coverage to 90.7\% while still requiring only 0.326 GFLOPs, or 31.6\% of the baseline.

On ADE20K-Object, adaptive stopping was evaluated using the fovea+context model together with canvas-based uncertainty. Since the low-resolution semantic prior is less reliable in realistic scenes, it is used only as a weak confidence cue, while object evidence is primarily accumulated from foveal observations. With $\alpha=0.5$, $\tau=0.4$, and a patience of 6, the model used 6.30 fixations on average, corresponding to 1.036 GFLOPs, or 23.8\% of the full-image baseline. At this setting, it reached 0.813 object recall, corresponding to approximately 4.31 of the 5.30 objects per image, with 0.480 Top-1 accuracy, 0.679 Top-3 accuracy, 0.078 full-canvas mIoU, and 0.194 visited-region mIoU. With $\alpha=0.5$, $\tau=0.3$, and a patience of 4, the model used 7.52 fixations on average, costing 1.228 GFLOPs, or 28.2\% of the baseline, and increased object recall to 0.836.

These results show that canvas confidence can be used to allocate computation adaptively: simple scenes require fewer fixations, while scenes with remaining salient uncertainty continue to receive additional observations.

\subsection{Goal-Directed Visual Search}

We additionally evaluated whether the semantic canvas can support a simple goal-directed visual search task without training a separate policy. Given a target class, the fixation score is biased toward regions where the current canvas assigns high probability to that class:

\begin{equation}
F_t(p)
=
\left(
A(p)
+
\alpha U_t(p)
+
\beta P_t(c_{\mathrm{target}} \mid p)
\right)
I_t(p),
\end{equation}

where $A(p)$ is bottom-up saliency, $U_t(p)$ is canvas uncertainty, $P_t(c_{\mathrm{target}} \mid p)$ denotes the probability assigned by the current semantic canvas to the target class at pixel $p$, obtained by applying a softmax over the accumulated canvas logits, and $I_t(p)$ is inhibition of return. The parameter $\beta$ controls the strength of the target bias.

As a preliminary goal-directed search experiment, we biased the fixation score toward regions where the semantic canvas predicted evidence for a requested target class. On ADE20K-Object, the strongest target-bias setting used 6.87 fixations on average, corresponding to 1.126 GFLOPs, or 25.8\% of the full-image baseline. The target class was observed before stopping in 77.9\% of images, compared with 76.1\% for the generic non-targeted policy at a similar threshold. The stricter criterion of fixating the target center was satisfied in 43.4\% of images. These results suggest that semantic confidence can weakly bias exploration toward task-relevant targets, but reliable goal-directed fixation remains an open direction.

These results suggest that the semantic canvas can support simple task-dependent exploration without learning a dedicated fixation policy. Importantly, the visual-search behavior is obtained by modifying the inference-time fixation score only; no reinforcement learning or task-specific policy training is introduced.

\section{Additional Qualitative Results}

\subsection{CLEVR Qualitative Results}
\vspace{-0.5em}

\begin{figure}[!htbp]
  \centering
  \includegraphics[width=\linewidth,height=0.62\textheight,keepaspectratio]{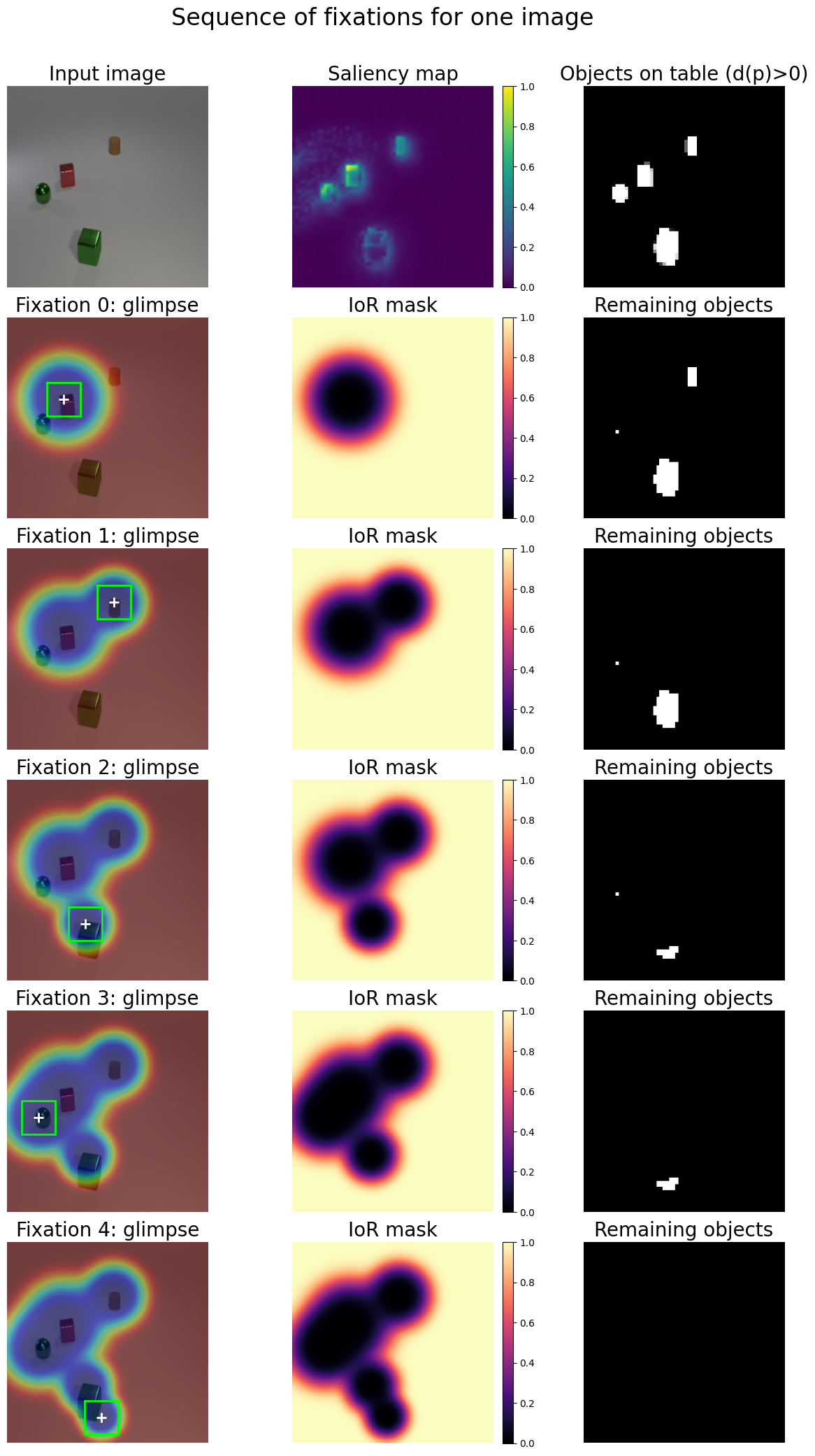}
  \caption{First row: input image (left), saliency map (middle), and objects on the table with depth $d(p)>0$ (right). Each subsequent row shows the current fixation region (left), the inhibited region (middle), and the objects remaining above the table (right).}
  \label{fig:supp_clevr_fixations}
\end{figure}

\begin{figure}[!htbp]
  \centering
  \includegraphics[width=0.9\linewidth,height=0.55\textheight,keepaspectratio]{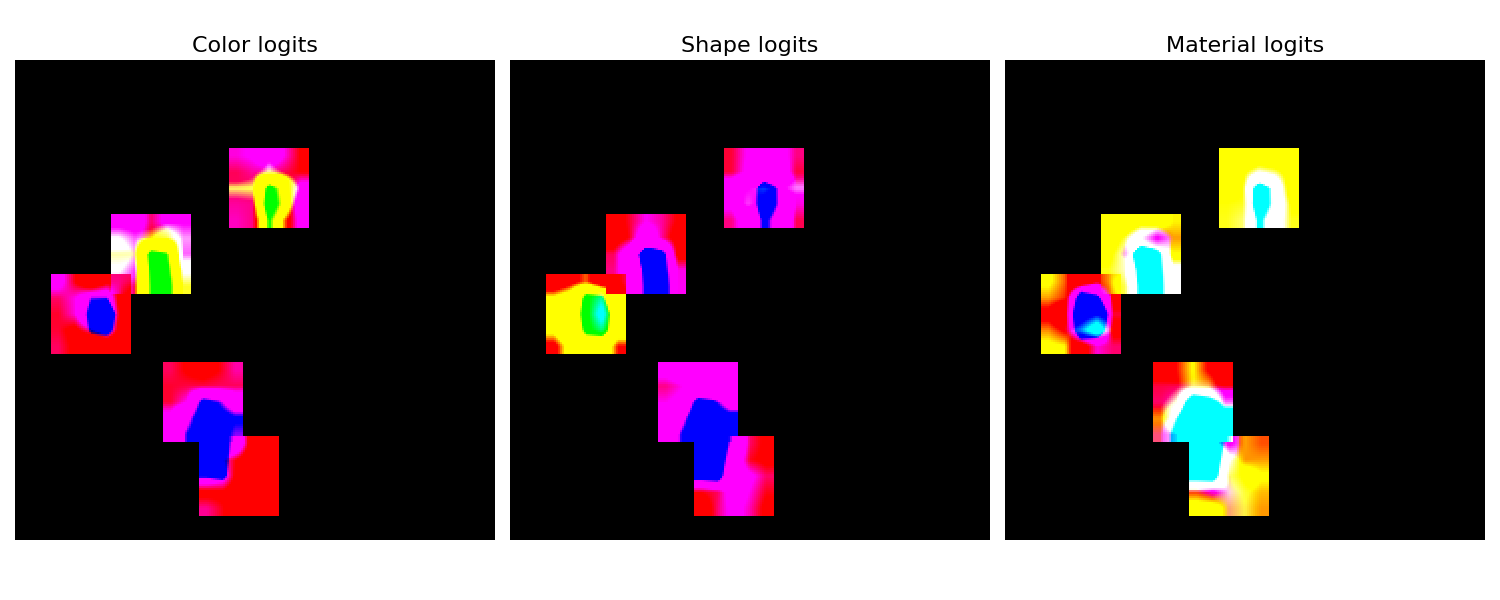}
  \caption{Reconstructed model prediction for the color head (left), shape head (middle), and material head (right). Different colors indicate different predicted classes the color, shape and material categories.}
  \label{fig:supp_clevr_predictions}
\end{figure}

Figures~\ref{fig:supp_clevr_fixations} and~\ref{fig:supp_clevr_predictions} provide qualitative examples of the sequential fixation process and the corresponding semantic predictions obtained from the progressively accumulated scene representation.
\FloatBarrier

\subsection{ADE20K Qualitative Results}
\vspace{-0.5em}

\begin{figure}[!htbp]
  \centering
  \includegraphics[width=0.9\linewidth,height=0.65\textheight,keepaspectratio]{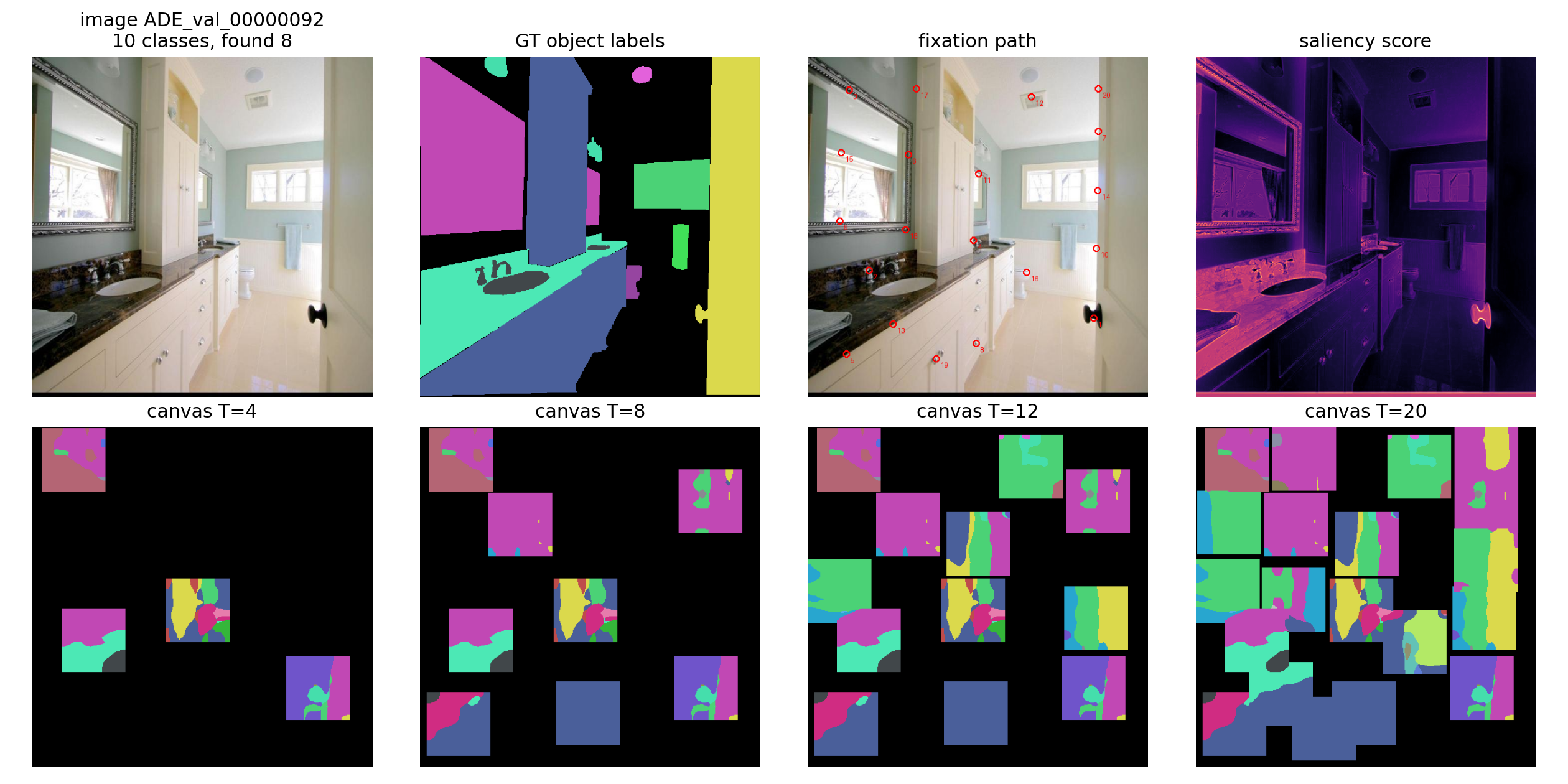}
  \caption{Qualitative ADE20K-Object example showing sequential semantic integration. The top row shows the input image, object labels, saliency-guided fixation path, and saliency map. The bottom row shows the semantic canvas after increasing numbers of fixations. As more foveal observations are accumulated, the canvas progressively covers more object regions and recovers a larger fraction of the scene.}
  \label{fig:supp_ade_canvas}
\end{figure}

Figure~\ref{fig:supp_ade_canvas} illustrates how the proposed semantic canvas progressively integrates semantic information from successive fixations.

\FloatBarrier

\end{document}